\documentclass[10pt]{article} 
\usepackage[accepted]{rlc}

\usepackage{amssymb}            
\usepackage{mathtools}          
\usepackage{mathrsfs}           
\usepackage{graphicx}           
\usepackage{subcaption}         
\usepackage[space]{grffile}     
\usepackage{url}                

\usepackage{wrapfig}
\usepackage[most]{tcolorbox}

\usepackage{indentfirst}
\usepackage{color}
\usepackage{amssymb}
\usepackage{fancyvrb}
\usepackage{pmboxdraw}
\usepackage{amsmath}
\usepackage{amsthm}
\usepackage{tikz}
\usepackage[hang,flushmargin]{footmisc}
\usepackage{chngcntr}
\usepackage{booktabs}
\usepackage{multirow}
\usepackage[makeroom]{cancel}
\usetikzlibrary{arrows,decorations.markings}

\usepackage{booktabs}
\usepackage{graphicx}
\usepackage{mathtools}
\usepackage{bbm}
\usepackage{relsize}
\let\oldnl\nl
\newcommand{\nonl}{\renewcommand{\nl}{\let\nl\oldnl}}
\makeatletter
\newcommand{\removelatexerror}{\let\@latex@error\@gobble}
\makeatother

\usepackage{enumitem}

\newenvironment{tightlist}%
{\begin{list}{$\bullet$}{%
    \setlength{\topsep}{0in}
    \setlength{\partopsep}{0in}
    \setlength{\itemsep}{0in}
    \setlength{\parsep}{0in}
    \setlength{\leftmargin}{1.5em}
    \setlength{\rightmargin}{0in}
}
}%
{\end{list}
}

\newcommand{\secref}[1]{Section \ref{#1}}

\newcommand{\figref}[1]{Figure~\ref{#1}}

\newcommand{\appref}[1]{Appendix~\ref{#1}}

\definecolor{green}{RGB}{0, 150, 0}

\counterwithin*{thm}{section}

\usepackage{mdframed}
\newmdtheoremenv[nobreak=true]{def2}{Definition}
\renewcommand{\cite}{\citep}

\def\thickhline{%
  \noalign{\ifnum0=`}\fi\hrule \@height \thickarrayrulewidth \futurelet
  \reserved@a\@xthickhline}
\def\@xthickhline{\ifx\reserved@a\thickhline
              \vskip\doublerulesep
              \vskip-\thickarrayrulewidth
             \fi
      \ifnum0=`{\fi}}

\newlength{\thickarrayrulewidth}
\DeclareUnicodeCharacter{225C}{$\triangleq$}
\DeclareUnicodeCharacter{2200}{$\forall$}
\newcommand{\wrt}{\textit{w.r.t.}}

\renewcommand{\S}{\mathcal{S}} 
\newcommand{\A}{\mathcal{A}}

\newcommand{\Ex}{\mathop{\mathbb{E}}}

\DeclareMathOperator*{\argmax}{argmax}
\DeclarePairedDelimiterX{\infdivx}[2]{(}{)}{%
  #1\;\delimsize\|\;#2%
}

\title{Sequential Decision-Making for Inline Text Autocomplete}

\author{Rohan Chitnis \\
  Meta AI \\\And
  Shentao Yang\thanks{\quad Work done while author was an intern at Meta AI. Correspondence to: \texttt{\{ronuchit, alborzg\}@meta.com}} \\
  UT Austin \\\And
  Alborz Geramifard \\
  Meta AI \\}

\begin{document}
\maketitle
\begin{abstract}
Autocomplete suggestions are fundamental to modern text entry systems, with applications in domains such as messaging and email composition. Typically, autocomplete suggestions are generated from a language model with a confidence threshold. However, this threshold does not directly take into account the cognitive load imposed on the user by surfacing suggestions, such as the effort to switch contexts from typing to reading the suggestion, and the time to decide whether to accept the suggestion. In this paper, we study the problem of improving inline autocomplete suggestions in text entry systems via a sequential decision-making formulation, and use reinforcement learning to learn suggestion policies through repeated interactions with a target user over time. This formulation allows us to factor cognitive load into the objective of training an autocomplete model, through a reward function based on text entry speed. We acquired theoretical and experimental evidence that, under certain objectives, the sequential decision-making formulation of the autocomplete problem provides a better suggestion policy than myopic single-step reasoning. However, aligning these objectives with real users requires further exploration. In particular, we hypothesize that the objectives under which sequential decision-making can improve autocomplete systems are not tailored solely to text entry speed, but more broadly to metrics such as user satisfaction and convenience.

\end{abstract}

\section{Introduction}
\label{sec:intro}

The ability to enter text is essential in today's world, spanning technology such as keyboards, phone/tablet screens, and smart watches. Autocomplete suggestions are fundamental to modern text entry systems, with applications in domains such as messaging and email composition. Autocomplete provides a mechanism for users to transmit more information without needing to enter too many more characters. For example, the user may simply enter ``\texttt{how a}'', upon which the autocomplete system suggests ``\texttt{how are you?}'', which can be accepted by the user in one additional stroke.

Among many possible instantiations, we choose to focus on \emph{inline} autocomplete systems, which may provide the user with up to a single suggestion at each timestep, displayed inline, e.g., ``\texttt{how a\color{gray}{re you?}}''. Such systems have been shown to be more effective at encouraging users to enter text than other approaches, such as displaying multiple suggestions in a different on-screen location~\cite{azzopardi2016analysis}. An inline autocomplete system must decide not only \emph{what} to suggest to the user, but also \emph{when} to make a suggestion to avoid interrupting users' focus~\cite{quinn2016cost}.

Existing autocomplete systems generally follow a two-stage process: 1) generate a ranked list of candidate completions of the current text entered by the user, 2) decide which, if any, to surface to the user based on a pre-defined fixed confidence threshold~\cite{cai2016survey,mitra2015query,quinn2016cost,fowler2015effects}. The downside of these systems is that they do not directly consider the cognitive load that surfacing autocomplete suggestions will impose on the user.
This may come from several aspects of the autocomplete process, such as switching contexts from typing to reading a suggestion, mentally processing the suggestion, and deciding whether to accept it.
Too many suggestions, even if usually accurate, can therefore lead to bad user experience.

\begin{figure}[t]
    \centering
    \includegraphics[width=0.49\columnwidth]{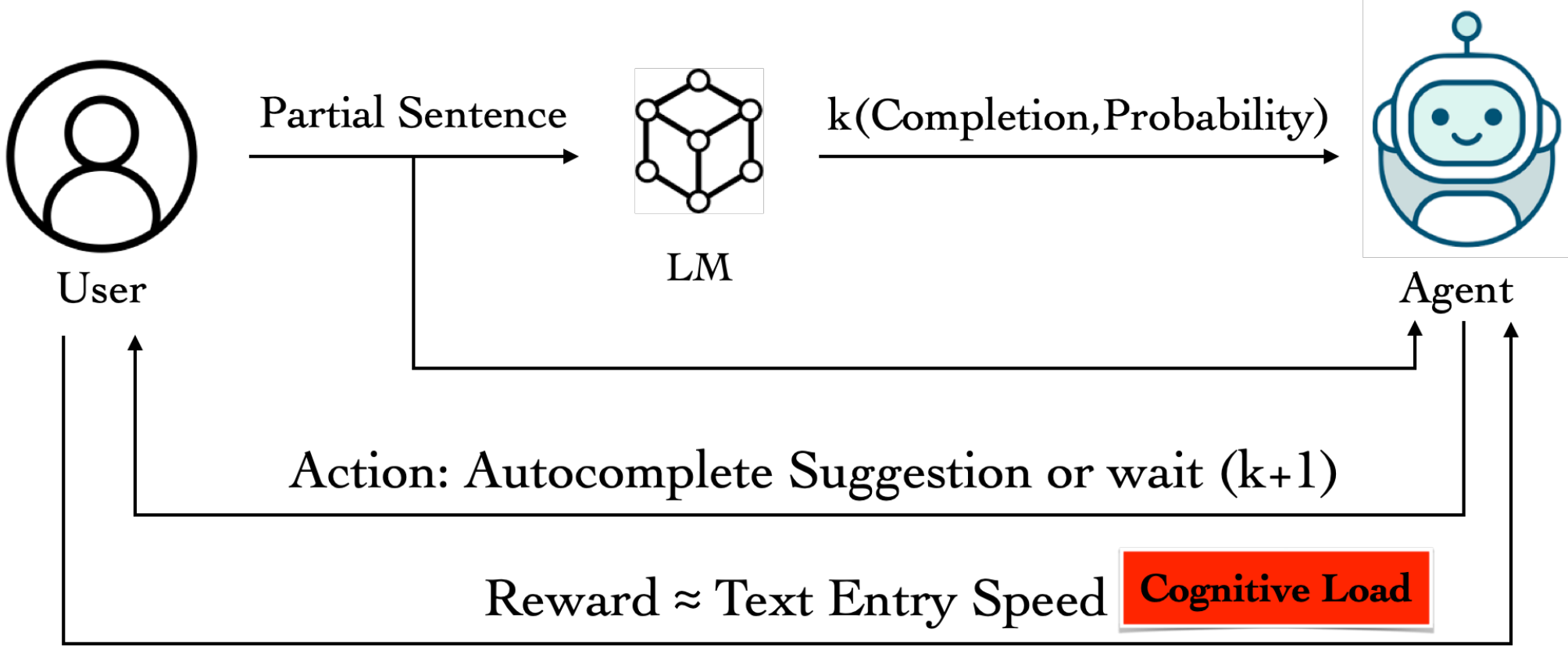}
    \includegraphics[width=0.49\columnwidth]{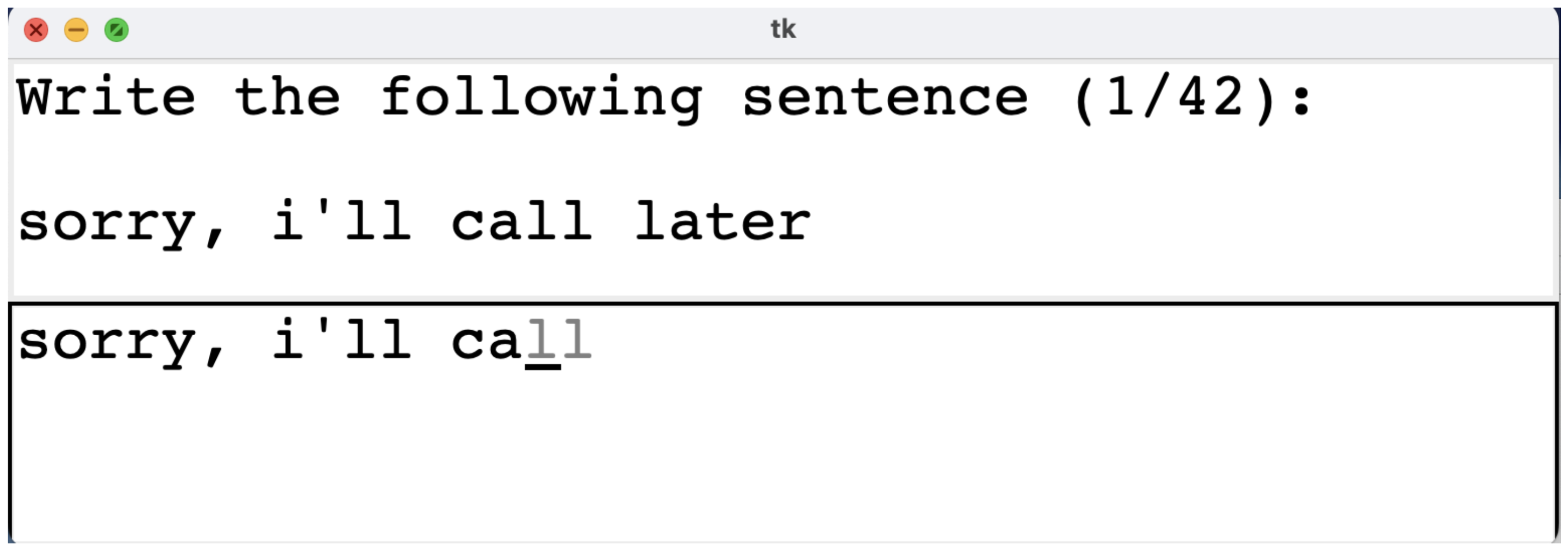}
    \caption{\emph{Left:} An overview of the workflow of our RL agent for inline text autocomplete. A language model (LM) generates $k$ candidate completions of the current text. The RL agent decides which, if any, to give the user as an inline suggestion. The agent is rewarded based on the user's text entry speed, which takes into account the cognitive load of showing suggestions. \emph{Right:} The interface for our user study (\secref{subsec:user_study}). In this example, the user was asked to type the sentence ``sorry, i'll call later'' on a keyboard. Currently, they have typed ``sorry, i'll ca'', and a suggestion was made that completes the last word as ``call'', which the user can accept by pressing the \texttt{tab} key.}
    \label{fig:teaser}
\end{figure}

We tackle the problem of generating inline autocomplete suggestions while minimizing the user's cognitive load by formulating text autocomplete as a sequential decision-making problem, and apply reinforcement learning (RL) methods~\cite{kaelbling1996reinforcement,sutton2018reinforcement} to train the autocomplete model. RL solves the problem of learning an optimal sequential decision-making policy through environment interaction, without assuming the differentiability of the objective/reward function \wrt\mbox{} the model parameters. 
RL methods have demonstrated impressive success in domains such as Atari~\cite{mnih2013playing} and Go~\cite{silver2016mastering}. 
Using RL methods to train an autocomplete model on top of a language model (LM) allows us to learn suggestion policies through repeated interactions with a target user, and define a reward function based on text entry speed that captures the cognitive load of surfacing suggestions (\figref{fig:teaser}).
We note that RL is suitable here since the cognitive load takes an unknown functional form \wrt\mbox{} the autocomplete model.

This paper's contributions are as follows:
\begin{tightlist}
    \item We formulated the inline text autocomplete problem as sequential decision-making.
    \item We performed theoretical analysis on when this formulation can improve over myopic reasoning.
    \item We ran simulated experiments on how an RL autocomplete agent performs on an idealized user.
    \item We performed a user study to understand the gaps between an idealized user and a real one.
\end{tightlist}

Our key findings are as follows:

(1) We acquired theoretical and experimental evidence that, when optimizing certain objective functions, RL can provide a better suggestion policy than myopic reasoning, but aligning these objectives with real-world users requires further exploration. We believe these objectives are not tailored solely to text entry speed, but more broadly to metrics like user satisfaction and convenience, which aligns with earlier findings in the text entry literature~\cite{quinn2016cost}.

(2) Given an idealized user that always accepts correct suggestions and inputs each character without typos, we could not find any evidence with a real dataset and  language model that RL-trained models improve text entry speed over the classical threshold-based approach with a fixed threshold value.

(3) Our user study ($N=9$) reveals that for keyboard typing, the cognitive load of looking at a suggestion is independent of its length, but is dependent on whether the suggestion matches what the user is trying to type (10ms if it matches, 50ms if it does not). Also, we could not find any evidence that suggestion acceptance rate is affected by the number of past surfaced suggestions.

\textbf{Recommendation.} Based on these findings, we recommend that further research into sequential decision-making for inline text autocomplete should \emph{not} pursue the goal of solely increasing text entry speed for idealized users. Instead, more realistic scenarios (stochastic user behavior, typos in their input, etc.), with a focus on user experience, may provide better opportunities for RL methods.

\section{Related Work}
\label{sec:rw}

Autocomplete has a rich history, especially in the domain of search queries~\cite{cai2016survey}, where methods that rely on neural language models have become increasingly popular~\cite{wang2020efficient,park2017neural,wang2018realtime}. For example, \citet{wang2020efficient} demonstrate how to make use of context in both the generation and ranking phases to improve end-to-end performance. These methods can also be used to obtain personalized models by embedding user IDs~\cite{fiorini2018personalized,jiang2018neural}. Like these works, we also use a neural language model in this paper for candidate suggestion generation, but for selection, we use a policy trained using deep reinforcement learning.

A separate line of work has investigated the empirical performance of existing autocomplete systems~\cite{fowler2015effects,quinn2016cost,azzopardi2016analysis}. \citet{fowler2015effects} found that the text entry enhancements found in modern touchscreen phones greatly reduce word error rate. \citet{quinn2016cost} discovered, surprisingly, that even though autocomplete suggestions can impair text entry speed, users often prefer them subjectively because they lowered cognitive and physical burden. \citet{azzopardi2016analysis} studied the cost-benefit tradeoffs that users make when conducting searches, concluding that inline autocomplete systems are an effective way to increase the amount of text entered. This entire line of work complements our paper, providing rationale for why inline autocompletion is a useful problem to study, and why cognitive load is important.

Reinforcement learning methods have demonstrated impressive success in domains such as Atari~\cite{mnih2013playing}, Go~\cite{silver2016mastering}, and control tasks~\cite{brockman2016openai}, but it is not commonly applied to text autocomplete systems. The two closest works to ours are \citet{wang2017learning} and \citet{lee2019learning}. The former studies query autocomplete as opposed to inline autocomplete; the authors formulated the problem as a multi-armed bandit~\cite{bandit1987}, a stateless simplification of the Markov decision process framework we adopt. The latter studied autocomplete as a two-agent communication game solved via unsupervised learning and optimized with policy gradients~\cite{sutton1999policy}. However, they assumed the user only enters keywords drawn from the target phrase, while we allow the user to input any English characters.

\section{Background: Reinforcement Learning}
\label{sec:background}

In this section, we give background for sequential decision-making and reinforcement learning in the Markov decision process (MDP) framework~\cite{puterman2014markov}.
An MDP is given by $\langle \S, \A, T, R, \gamma \rangle$, with state space $\S$; action space $\A$; transition model $T(s_t, a_t, s_{t+1}) = P(S_{t+1}=s_{t+1} \mid S_{t}=s_t, A_t=a_t)$, where $t$ is the time-step, $s_t, s_{t+1} \in \S$, $a_t \in \A$, and $S_t$, $A_t$, $S_{t+1}$ are random variables; reward function $R(s_t, a_t, s_{t+1}) = r_t \in \mathbb{R}$; and discount factor $\gamma \in [0, 1]$.
The optimal solution to an MDP is a policy $\pi^*: \S \to \A$, a mapping from states to actions, such that acting under $\pi^*$ maximizes \emph{return}, the expected sum of discounted rewards: $\pi^* = \argmax_\pi \Ex\left[\sum_{t=0}^H \gamma^t R\left(s_t, \pi(s_t), s_{t+1}\right)\right]$. Here, $H$ is the horizon of an episode, a sequence of state-action-rewards from an initial to a terminal state. The optimal $Q$-value of a state-action pair, $Q^*(s_t, a_t)$, is the expected return of taking action $a_t$ from state $s_t$, and acting optimally afterward: $Q^*(s_t, a_t) = \Ex_{s_{t+1} \sim T(s_t, a_t, \cdot)} \left[R(s_t, a_t, s_{t+1}) + \gamma  V^*(s_{t+1}) \right]$. The optimal value of a state, $V^*(s_t)$, is the maximum $Q$-value over actions: $V^*(s_t) = \max_a Q^*(s_t, a)$.

In reinforcement learning, an agent interacts with an MDP where $T$ and $R$ are unknown, and attempts to discover $\pi^*$ through these interactions. There are many families of algorithms for RL~\cite{kaelbling1996reinforcement,arulkumaran2017brief,sutton2018reinforcement}, but some of the most popular are policy gradient methods and value-based methods. The former optimize a policy directly~\cite{sutton1999policy,schulman2017proximal,schulman2015trust}, while the latter learn the value of each state-action pair and use that to infer the policy~\cite{mnih2013playing,watkins1992q}. Another recently popularized direction is offline RL~\cite{levine2020offline}, in which the agent does not get to interact with the environment, but rather must learn from a static dataset of prior interactions.

\section{Problem Formulation: Autocomplete as Sequential Decision-Making}
\label{sec:formulation}

In this section, we describe how to formulate inline text autocomplete as an MDP, which can then be solved via RL. As shown in \figref{fig:teaser}, we assume access to a language model (LM) that can generate $k$ candidate completions of a partial sentence, along with their probabilities. We build the MDP on top of this LM.
A state in the MDP is the current \emph{context} and the $k$ candidates from the LM (with their probabilities). The context is all characters entered so far in the sentence, which includes the full previous words and the prefix of the current word. The action space has size $k+1$: surfacing one of the $k$ candidates to the user, or a special \texttt{wait} action that does not surface anything. Note that in this formulation, the word corresponding to each action $a \in \{1,\ldots, k\}$ (except for \texttt{wait}) changes on each step based on the LM output. While the agent should do better with higher $k$ since it has more candidates to choose from, increasing $k$ would also increase the problem complexity.

The transition model is defined by the user and LM. On each timestep, after the RL agent acts, the user enters a character. The character can be a special acceptance key (e.g., \texttt{tab}), or the next English character the user wants to enter. After that, the LM generates $k$ new candidates from the updated context. In our computational experiments (\secref{subsec:computational}), we will consider an \emph{idealized user} who behaves as follows. They sample a target sentence at the start of each episode, unknown to the RL agent. On each timestep, the idealized user accepts the suggestion if and only if it matches any prefix of the remaining sentence. In case of no suggestion or a mismatched suggestion, the idealized user enters the next English character of the current word without making any typos.\footnote{This transition model is stochastic from the perspective of the RL agent because it does not know the target sentence. We could alternatively model this using a partially observable MDP~\cite{kaelbling1998planning}, but we stick with the fully observable formulation due to its simplicity and popularity in the RL field.} An episode ends when all characters of the target sentence have been entered.

Finally, the reward function is proportional to the time saved/lost due to the suggestions:
\begin{equation*}
R = \begin{cases}
0 & \text{\small{no suggestion made by agent}}\\
(1-\alpha) \cdot \text{len(suggestion)} - \beta &  \text{\small{suggestion accepted by user}}\\
-\alpha \cdot \text{len(suggestion)} - \beta &  \text{\small{suggestion ignored by user}}
\end{cases} \,.
\end{equation*}
Here, $\alpha,\beta \in [0, 1]$ are parameters controlling the degree of penalty on the agent when it makes a suggestion, due to the user's cognitive load. $\alpha$ is the cognitive load proportional to suggestion length, while  $\beta$ is a constant cognitive load incurred for moving the gaze.
Refer to \appref{sec:appendix:reward} for the derivation of the reward function.
By default, we use $\alpha = 40/521$ (character reading time / character writing time) and $\beta = 60/521$ ($2 \times$ saccade time / character writing time). If the user is idealized, then under this reward function, \texttt{wait} actions give zero reward, correct suggestions accepted by the user give positive reward, and incorrect suggestions ignored by the user give negative reward. The choice of $\gamma$ can heavily influence the optimal solution; our experiments will explore this further.

\section{Theoretical Analysis}
\label{sec:theory}

We begin with a theoretical study of a simple setting where a user may be entering one of two words with equal probability. Our goal is to answer: \emph{Do there exist conditions such that the sequential decision-making formulation of text autocomplete is beneficial over myopic reasoning?} 
To answer this question, we study a restricted setup and derive conditions on the reward function under which the optimal policy in the MDP with $\gamma=1$ obtains higher return at a particular timestep than the optimal (myopic) policy in the MDP with $\gamma=0$, holding all other components fixed. In particular, we aim to show that the farsighted agent \emph{waits} for more information in certain cases where the myopic agent would make an incorrect suggestion, which gives the farsighted agent higher return.

\subsection{Analysis Setup}
\label{subsec:analysis_setup}
We focus on an idealized user (\secref{sec:formulation}) and $k=1$, meaning there are two actions to choose from at each timestep: \textit{show} the top LM candidate suggestion ($S$) or \textit{wait} ($W$).
To keep the analysis simple, we consider two arbitrary words $u$ and $v$ of length $n$. Both words share the first $m$ letters, e.g., for words ``this'' and ``they'' we have $n=4$, $m=2$. The user picks the target word uniformly at random across both at the start of each episode. We aim to find $\alpha$, $\beta$ in our reward function such that the optimal policy for $\gamma=1$ waits while the optimal policy for $\gamma=0$ shows, at the last common letter (position $m$).
Note that this is sufficient to answer our central question stated above, because the $Q$-value of the $\gamma=1$ policy is precisely the sum of future rewards, our evaluation criteria. Due to greedy action selection,  the $\gamma=1$ policy will find a better action than the $\gamma=0$ policy, should they behave differently.




We consider a specific form of the LM that outputs the non-target word as the candidate when $\leq m$ letters are entered, and the target word after that.
Let $u_t$ be the state of the $t^{\mathrm{th}}$ step consisting of the first $t$ letters of $u$, and  similarly for $v_t$, for any $t \leq n$.  For notational simplicity, we use ``$Q(u_t, \cdot)$'' to refer to the optimal $Q$-value of the state $u_t$ (similarly with $v_t$) and action either $S$ or $W$. Note that this notation hides the LM candidates portion of the state.

\subsection{Derivation for $\gamma=0$}
\label{subsec:gamma0deriv}
We begin by deriving conditions on $\alpha$ and $\beta$ under which the optimal policy for $\gamma=0$ would prefer to suggest ($S$) at $t=m$. The base case is
$Q(u_n, W) = Q(u_n, S) = Q(v_n, W) = Q(v_n, S) = 0$. Then:
\begin{align*}
\hline
\text{Case 1: } m &< t \leq n \\
\hline
Q(u_t, W) = Q(v_t, W) &= 0\\
Q(u_t, S) = Q(v_t, S) &= (n-t)(1-\alpha)-\beta\\
\hline
\text{Case 2: } t &\leq m \\
\hline
Q(u_t, W) = Q(v_t, W) &= 0\\
Q(u_t, S) = Q(v_t, S) &= 0.5[(n-t)(1-\alpha)-\beta]+0.5[-(n-t)\alpha-\beta]\\ &= (n-t)(0.5-\alpha)-\beta \\
\hline
\end{align*}

The condition for preferring suggesting ($S$) at $t=m$ under $\gamma=0$ is $Q(u_m, S) > Q(u_m, W)$, which implies that $(n-m)(0.5-\alpha)-\beta > 0$.

\subsection{Derivation for $\gamma=1$}
\label{subsec:gamma1deriv}
Now, we derive conditions on $\alpha$ and $\beta$ under which the optimal policy for $\gamma=1$ would prefer to wait ($W$) at $t=m$. This derivation is more involved as the $Q$-value expressions are recursive, due to lookahead. Again, the base case is
$Q(u_n, W) = Q(u_n, S) = Q(v_n, W) = Q(v_n, S) = 0$. Now:
\begin{align*}
\hline
\text{Case 1: } m &< t \leq n \\
\hline
Q(u_t, W) &= \max(Q(u_{t+1}, W), Q(u_{t+1}, S))\\
Q(v_t, W) &= \max(Q(v_{t+1}, W), Q(v_{t+1}, S))\\
Q(u_t, S) &= Q(v_t, S) = (n-t)(1-\alpha)-\beta\\
\hline
\text{Case 2: } t &\leq m \\
\hline
Q(u_t, W) &= Q(v_t, W) = 0.5\max(Q(u_{t+1}, W), Q(u_{t+1}, S))+0.5\max(Q(v_{t+1}, W), Q(v_{t+1}, S))\\
Q(u_t, S) &= Q(v_t, S) = 0.5[(n-t)(1-\alpha)-\beta]+0.5[-(n-t)\alpha-\beta+\max(Q(v_{t+1}, W), Q(v_{t+1}, S))]\\
\hline
\end{align*}

With some algebra, we can solve the recursion for the $t=m$ case that we care about:
\begin{align*}
Q(u_m, W) &= (n-m-1)(1-\alpha)-\beta\\
Q(u_m, S) &= (n-m)(1-1.5\alpha)-0.5(1-\alpha)-1.5\beta\\
\end{align*}

The condition for preferring wait ($W$) at $t=m$ under $\gamma=1$ is $Q(u_m, S) < Q(u_m, W)$, which implies that $(n-m+1)\alpha+\beta > 1$.

\subsection{Final Constraints on $\alpha$ and $\beta$}
Putting together the work in \secref{subsec:gamma0deriv} and \secref{subsec:gamma1deriv}, we finally obtain the following constraint on $\alpha$ and $\beta$ which would lead to different optimal policies for $\gamma=0$ and $\gamma=1$ at at position $m$:
\begin{align*}
\Aboxed{
    \frac{1-\beta}{n-m+1} < \alpha < \frac{0.5(n-m) -\beta}{n-m}
} \;.
\end{align*}

With a fixed value of $\beta$, an $\alpha$ value close to 0.5 can satisfy the above. To go beyond the two-word setup empirically, we obtained the most likely 500 words from our LM (described in \secref{subsec:exp_setup}), calculated the optimal policy using backward dynamic programming for $\gamma=0$ and $\gamma=1$ (other MDP components are the same), and measured the number of states for which the two policies differed. \figref{fig:diff_vs_alpha} below depicts this difference for various $\alpha$. We set $\beta$ to the default value $60/521$.

\begin{wrapfigure}{R}{0.51\columnwidth}
    \centering
\includegraphics[width=0.5\columnwidth]{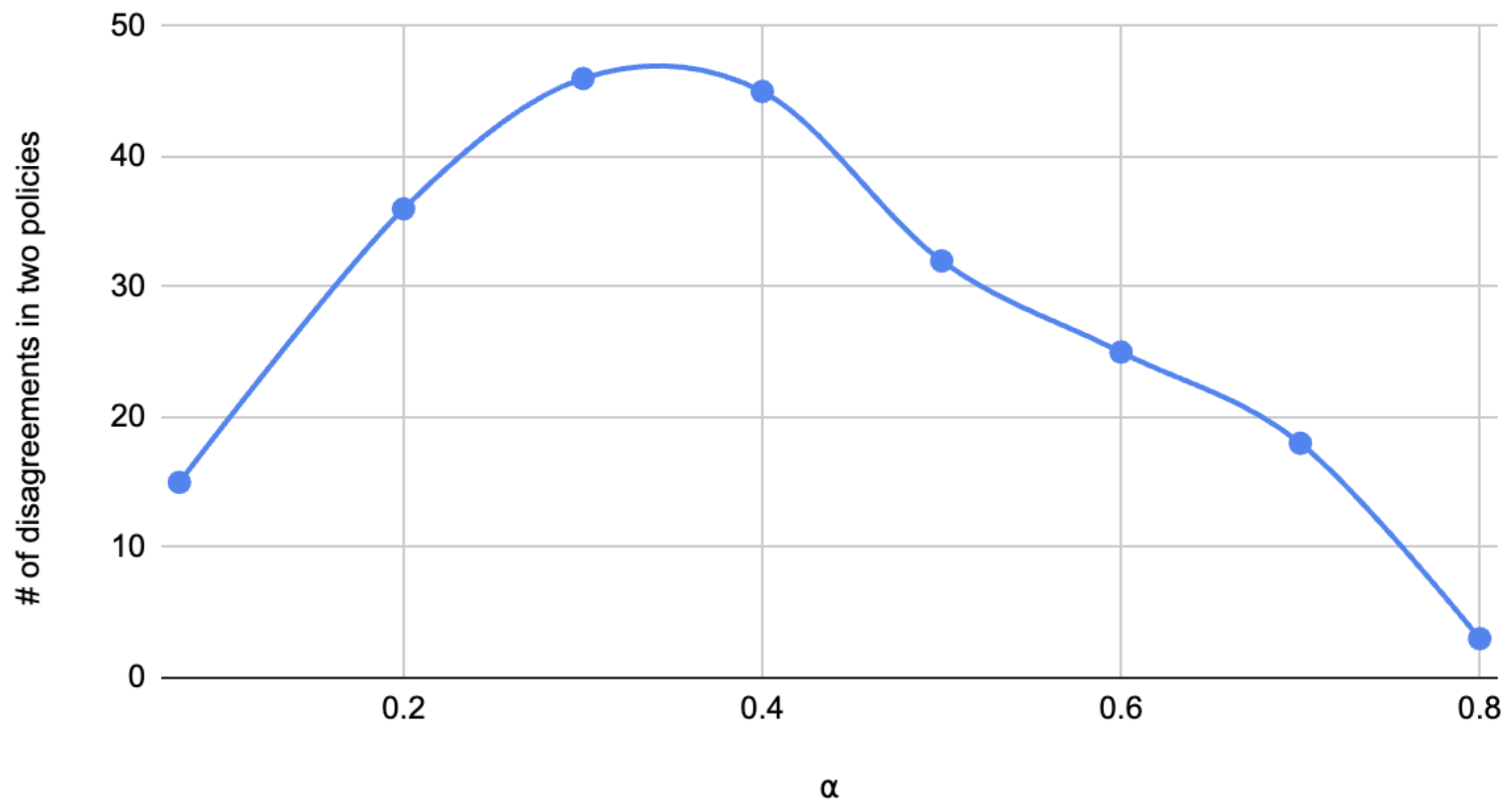}
    \vspace{-1em}
    \caption{Number of states where the optimal farsighted policy ($\gamma=1$) and the optimal myopic policy ($\gamma=0$) disagree, for various values of $\alpha$.}
    \label{fig:diff_vs_alpha}
\end{wrapfigure}

We can see that the gap is largest when $\alpha$ is between 0.3 and 0.4. Hence, for these values of $\alpha$, we should see more benefit to solving the autocomplete problem via the sequential decision-making formulation; however, the disagreement ($y$-axis) only corresponds to $\sim$2\% of the state space. Another intuition for using larger $\alpha$ values is to more strongly penalize long but incorrect suggestions, which incentivizes an RL agent to wait for additional input from the user before being informed enough to make a suggestion. Notice that as $\alpha$ gets closer to 1, both policies become very conservative due to the high potential penalty of making wrong suggestions, and hence the number of disagreements reduces drastically.

\subsection{Limitations of our Analysis}
Although we derived constraints under which text autocomplete benefits from sequential reasoning, these constraints may not reflect the goal of text entry speed improvement. In our user study (\secref{subsec:user_study}), we will see that the ideal values of $\alpha$ and $\beta$ for faster text entry are near zero, which violates the constraints we derived.
Additionally, our analysis depends on a simple form of LM which is unlikely to reflect those used in practice. The degree to which the sequential decision-making formulation can benefit our problem depends on the user's goal and what LM we use to implement the MDP.

\section{Experiments}
\label{sec:experiments}

\subsection{Experimental Setup}
\label{subsec:exp_setup}
We sampled target sentences for the user from two open-source datasets, SMS Spam Collection (``ham'' labels only)~\cite{almeida2011contributions} and Reddit Webis-TLDR-17~\cite{volske-etal-2017-tl}. Applying the following filters led to $\sim$500 and $\sim$300,000 sentences in each dataset respectively:
\begin{tightlist}
    \item Use only sentences with length $\leq 10$ words to save compute.
    \item Use only sentences with no profanity and whose perplexity is under 150 as computed by GPT2-Large \cite{gpt2}, to ensure language quality (e.g., no misspellings or slang).
    \item Remove sentences with characters not in the set ``a-z,.’?! '' to satisfy the LM input requirements.
\end{tightlist}

For our language model (LM), we trained a small ($\sim$25MB) transformer~\cite{vaswani2017attention} on the Wikipedia\footnote{\url{https://huggingface.co/datasets/wikipedia}} and Reddit Webis-TLDR-17 corpora. This transformer outputs a probability distribution over the English vocabulary. To reduce down to the $k$ candidates fed into our RL agent, we take the top $k$ outputs and renormalize their probabilities. Unless otherwise specified, our experiments only consider single-word autocompletion (as opposed to multi-word). 

For all simulations, we assume the user is the idealized one described in \secref{sec:formulation}.

\subsection{Computational Experiments}
\label{subsec:computational}

We explored three RL algorithms: PPO, an online policy gradient method~\cite{schulman2017proximal}; DQN, an online value-based method~\cite{mnih2013playing}; and IQL, an offline method~\cite{kostrikov2021offline}. We benchmark these algorithms against three baseline agents: oracle, uniform random, and threshold-based. The oracle agent has privileged access to the target sentence and always surfaces correct suggestions when given by the LM; it thus establishes an upper bound on performance. The threshold-based agent surfaces the top suggestion if and only if its probability is above a fixed threshold. Optimization on the validation set led us to use threshold=$0.3$. For PPO and DQN, we used DistilBERT~\cite{sanh2019distilbert} as the policy architecture and initialization, while IQL used a multi-layer perceptron (MLP) trained from scratch. See \appref{app:exp_details} for more details.

The IQL policy received the suggestion probabilities from the LM as part of its input, while PPO/DQN policies did not. This design choice was made because probabilities from the LM are floating-point numbers and it is not useful to tokenize them as strings to feed into transformers. Future work may consider learning or constructing an input representation for these probabilities, e.g., using an embedding layer~\cite{chen2021decision} or predefined activation~\cite{gorishniy2022embeddings}.

All RL algorithms were trained for 250K gradient steps with $\gamma=0.99$. Since IQL is an offline RL algorithm, we trained it on a static dataset of 5,000 trajectories collected by a policy that follows the threshold-based baseline but with a 5\% chance of taking a random exploratory action each step.

\begin{figure}[t]
    \centering
    \includegraphics[width=0.49\columnwidth]{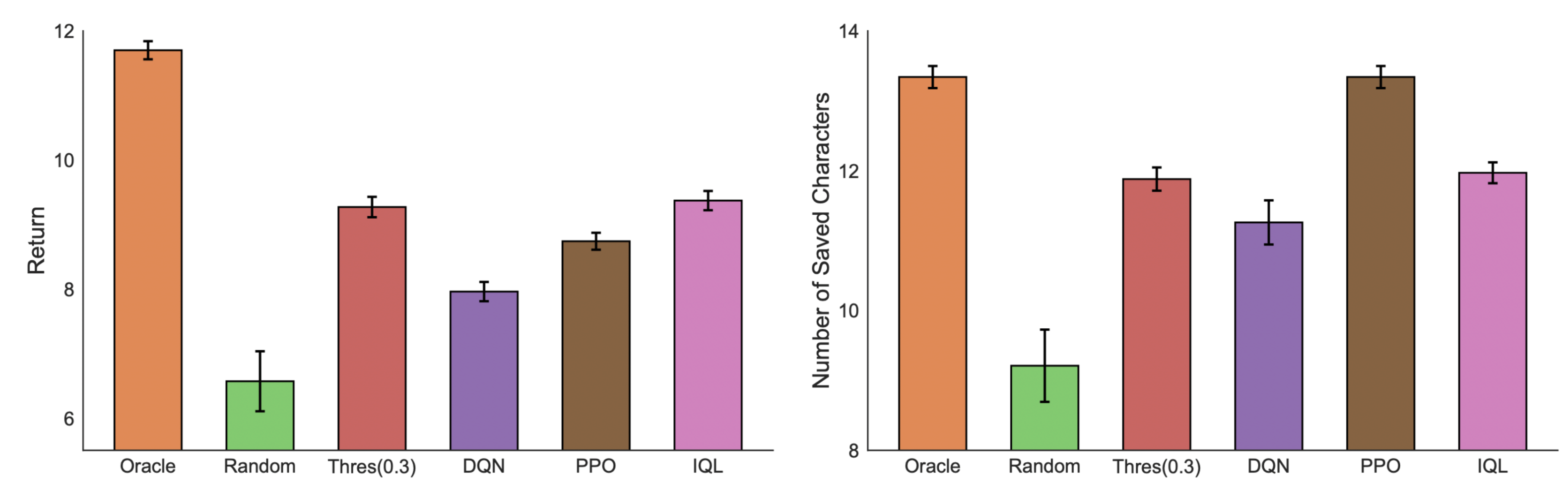}
    \includegraphics[width=0.49\columnwidth]{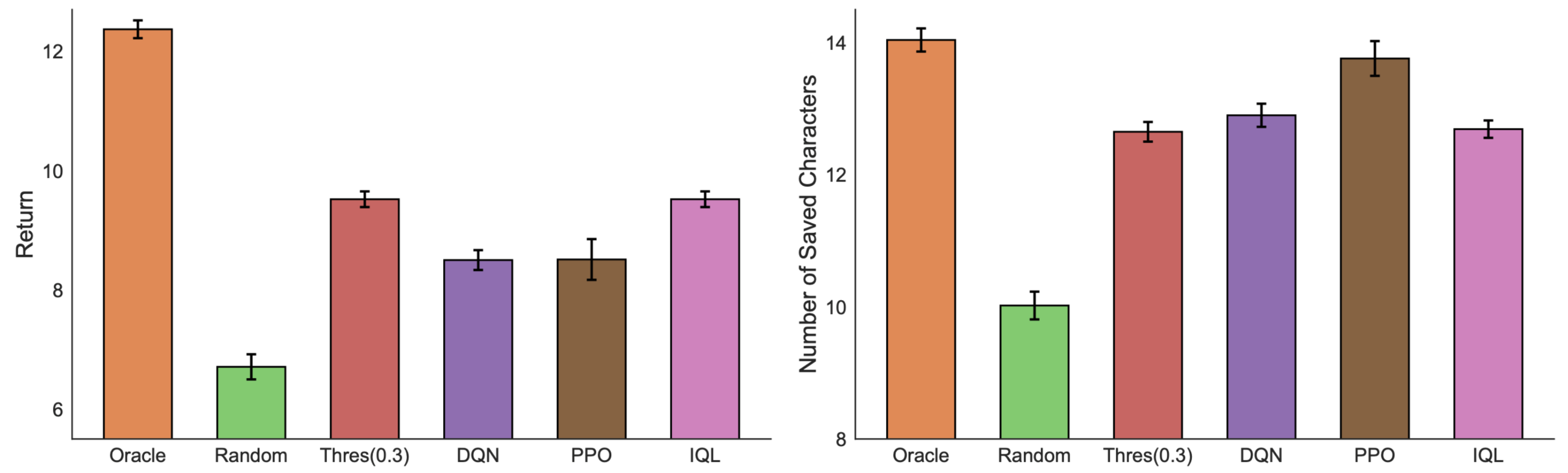}
    \caption{Average return and number of characters saved, over 5 independent runs (including training, for RL agents). Error bars depict 95\% confidence interval. \emph{Left two:} SMS dataset, 50 sentences in evaluation set. \emph{Right two:} Reddit Webis-TLDR-17 dataset, 400 sentences in evaluation set.}
    \label{fig:comp_baseline}
\end{figure}

First, we considered the $k=1$ setting, where the LM generates one candidate and the action space is $\{suggest, wait\}$. We used the default reward function parameters $\alpha=40/521$, $\beta=60/521$. \figref{fig:comp_baseline} shows the average return and number of saved characters in our two domains.
In both cases, IQL performed best among RL techniques (9.37 $\pm$ 0.17 average return), but could not improve significantly over the threshold-based agent (9.27 $\pm$ 0.18). Interestingly, PPO was overly aggressive: it saved the user the most characters, yet it obtained a lower return due to more incorrect suggestions.

\begin{figure}[t]
    \centering
    \includegraphics[width=0.49\columnwidth]{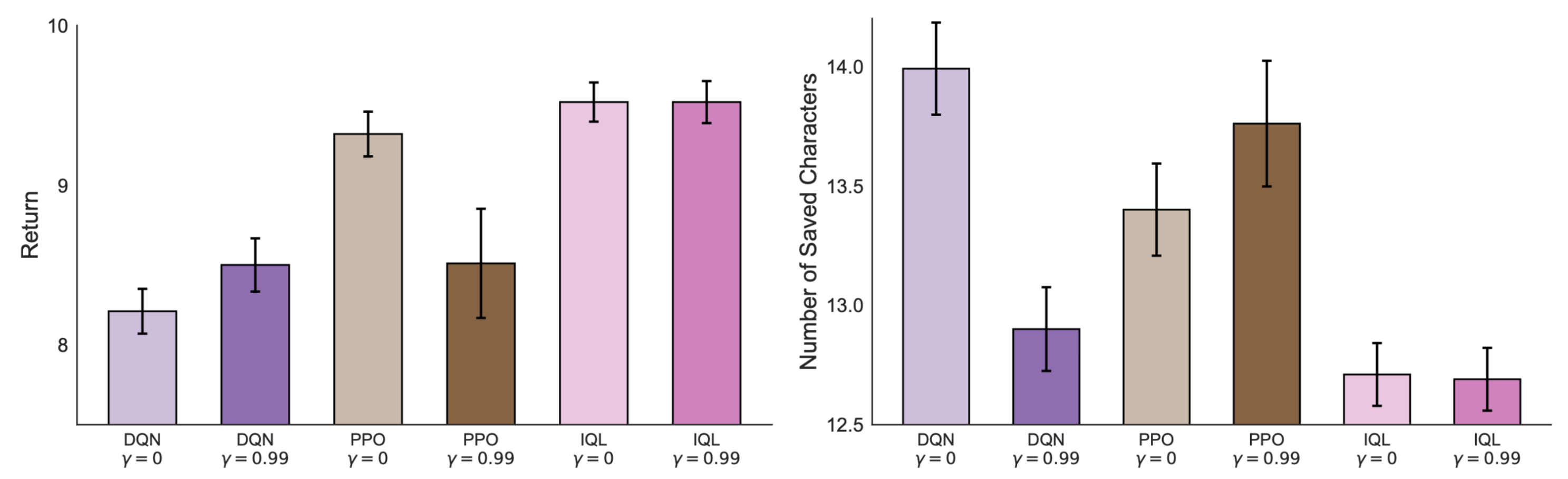}
    \includegraphics[width=0.49\columnwidth]{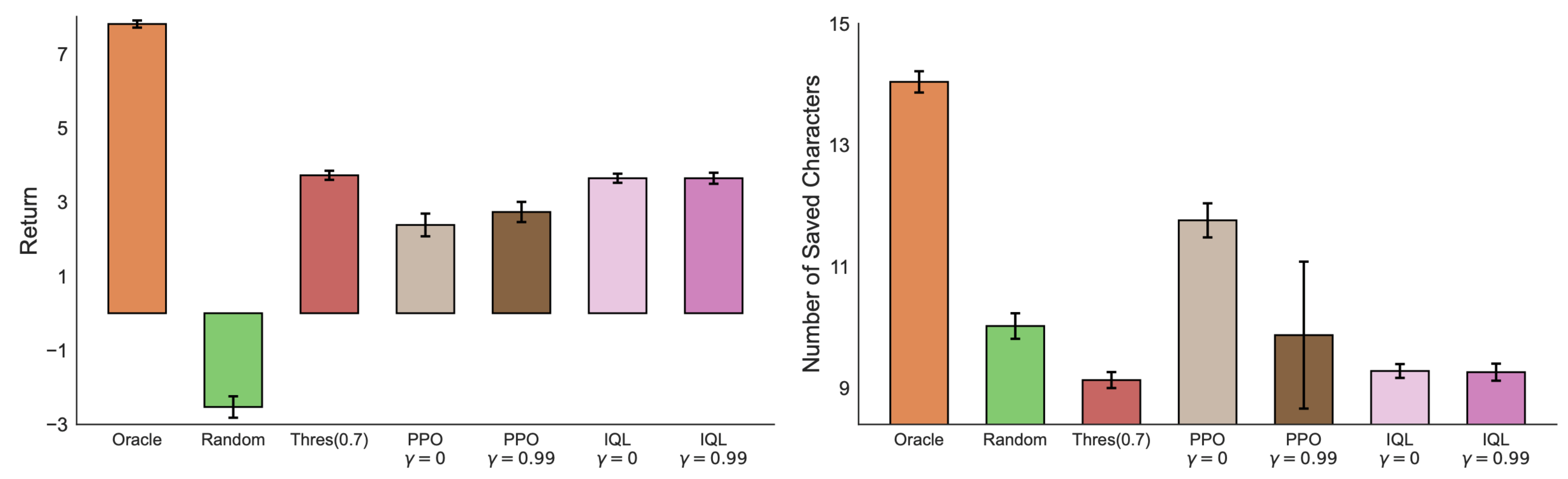}
    \caption{\emph{Left two plots:} Analysis of how varying $\gamma$ affects RL agents. \emph{Right two plots:} Results of rerunning experiments with $\alpha=0.4$, as suggested by our theory (\secref{sec:theory}). All plots use the Reddit Webis-TLDR-17 dataset and show the same metrics (y-axis) as in \figref{fig:comp_baseline}. In the right two plots, DQN results are omitted because they were significantly worse than PPO and IQL, and we changed the threshold-based agent to use threshold 0.7 based on re-tuning with the updated $\alpha$.}
    \label{fig:comp_additional}
\end{figure}

These observations made us question whether our autocomplete problem benefits from sequential decision-making, or if a contextual bandit suffices. To answer this, we reran the RL agents under the same MDP but with $\gamma=0$. Results are shown in \figref{fig:comp_additional} (left two plots). Theoretically, $\gamma=0.99$ renders the problem harder, yet the solution asymptotically should be better or the same as when $\gamma=0$. For IQL, the results with both $\gamma$ values were on par. However, for PPO, we observed better results with $\gamma=0$, though as we increased the number of training steps to 1M, the difference disappeared (not shown in plots).
Our observations support the hypothesis that with our current setup, sequential decision-making may not be helpful.

Next, we incorporated our theoretical takeaways from \secref{sec:theory} and set $\alpha=0.4$ to see if this provides better opportunity for RL. Results are shown in \figref{fig:comp_additional} (right two plots).
Indeed, our theoretical analysis transferred to PPO: we see that the PPO agent trained under $\gamma = 0.99$ performs better than $\gamma = 0$. Because this difference was not significant after 250K steps, we extended the learning to 1M steps and confirmed the returns differed significantly (not shown in plots): (3.14 $\pm$ 0.27) for $\gamma = 0.99$ vs. (2.65 $\pm$ 0.16) for $\gamma = 0$. However, like in \figref{fig:comp_baseline}, PPO could not reach the average return of IQL and the threshold-based agent. We believe this gap in performance is due to the fact that the PPO policy does not receive suggestion probabilities from the LM as part of its input.


Our final computational experiment revolved around \emph{multi-word suggestions}, drawing on the intuition that longer suggestions have more opportunity to improve text entry speed. We increased our candidate pool to $k=5$ and allowed LM candidates to be 1 or 2 words. Since the joint likelihood of suggesting two words is always lower than the likelihood of suggesting just the first, we normalized the joint probability of 2-word suggestions, similar to \citet{murray2018correcting}. See \appref{sec:appendix:len_norm} for details. Unfortunately, our initial experiments indicated that allowing multi-word suggestions degraded the return for both oracle and threshold-based agents. Surprisingly, it also reduced the oracle agent's number of saved characters. The reason is that 2-word candidates can take up the slot of the correct 1-word candidate, preventing it from being seen by the agent. See \appref{sec:appendix:multiword_bad} for details. Given these findings, we didn’t train RL agents (which are GPU-intensive) in this setting.

\subsection{User Study}
\label{subsec:user_study}
In addition to our computational experiments, we ran a user study to understand how well our reward function aligns with real user behavior.
Our goals were to 1) estimate real values for the reward function parameters, $\alpha$ and $\beta$, that capture the cognitive load of the user, and 2) explore the possibility of ``accumulated fatigue''~\cite{Paas_2003} for the users, meaning that as the number of surfaced suggestions increases, the user is less likely to accept suggestions.
To address these goals, we ran a study where we asked $N=9$ users to enter 42 sentences on a keyboard, both with and without autocomplete suggestions. \figref{fig:teaser} (right) shows an example of the study interface. All $N=9$ users work with computers as a regular part of their profession.

\begin{figure*}
    \centering
    \includegraphics[width=0.40\textwidth, height= 4.5cm]{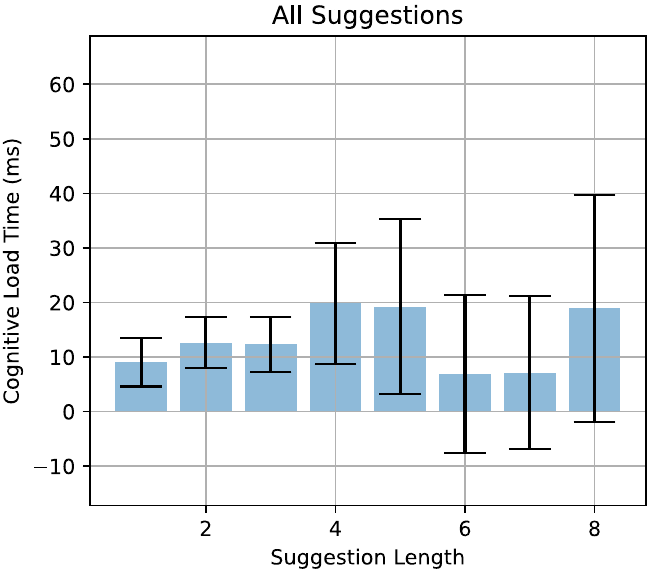}
    \includegraphics[width=0.42\textwidth, height= 4.4cm]{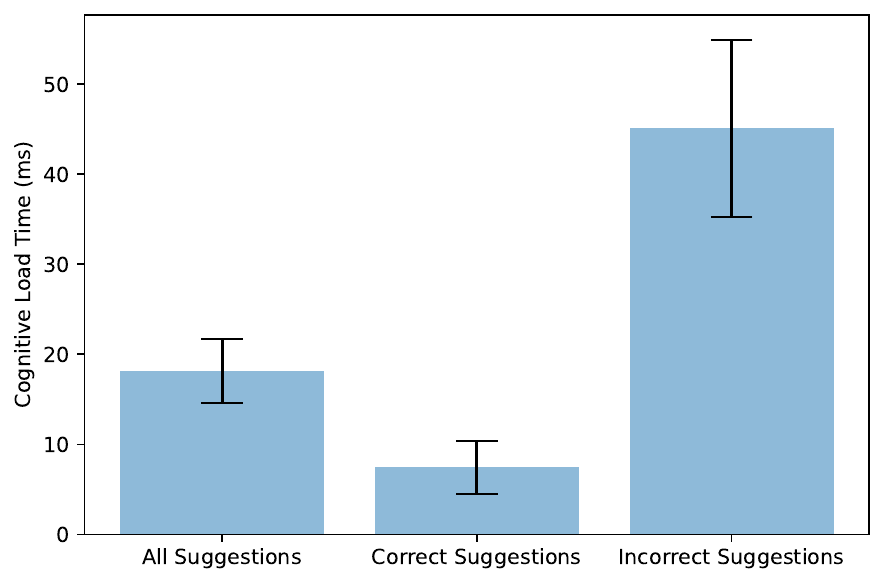}
    \caption{User study results (\secref{subsec:user_study}). Error bars depict $95\%$ confidence interval. \emph{Left:} Average cognitive load versus suggestion length. The cognitive load did not grow significantly with suggestion length. \emph{Right:} Average cognitive load versus suggestion correctness. There is a significant difference between the user's cognitive load when considering correct versus incorrect suggestions.
    }
    \label{fig:user_study_results}
\end{figure*}

We obtained 3968 instances where a user entered the same key given the same context (letters typed so far) both with and without autocomplete suggestions. To measure the cognitive load due to the suggestions, we looked at the time difference between how long it took the user to enter the next key in both cases. \figref{fig:user_study_results} (left) shows this cognitive load measure as a function of suggestion length. Contrary to our initial estimate of $\alpha=40/521$, the time users spend on suggestions did not grow with suggestion length, implying that $\alpha$ should be set to 0. The average cognitive load for users was $21.29 \pm 3.61$ms, which falls within the 20-30ms of a single saccade time for reading~\cite{Rayner2009-rm}. This observation suggests that the user only has to saccade once, not twice, after reading a suggestion. This is likely because initially, their eyes are already focused where the suggestion appears, which can in return explain the widespread adoption of inline suggestions.

Furthermore, \figref{fig:user_study_results} (right) shows that the cognitive load of looking at a suggestion is highly dependent on suggestion correctness.  Specifically, users experienced cognitive loads of $9.18 \pm 3.05$ms and $50.49 \pm 9.67$ms for correct and incorrect suggestions respectively. This observation suggests that $\beta$ is dependent on suggestion correctness, as opposed to our initial fixed estimate of $\beta=60/521$.

Finally, we probed the presence of ``accumulated fatigue'', as defined earlier. See \figref{fig:appendix:fatigue_results} in \appref{sec:appendix:cog-load} for results. The data did not show any evidence of declined acceptance rate based on the cumulative number of past suggestions, or the cumulative number of past incorrect suggestion.

Based on these findings, we re-ran our experiments on the Reddit Webis-TLDR-17 dataset with $\alpha=0$, and $\beta = 10/521$ if the suggestion is correct and $50/521$ if it is wrong. With these changes, we found that the optimal threshold for the threshold-based agent was 0: always surface the top LM candidate, regardless of its probability. This is because by reducing $\alpha$ and $\beta$, we reduced the penalty for incorrect suggestions. This result implies that the sequential decision-making formulation does \emph{not} improve text entry speed of an idealized user when $\alpha, \beta$ are chosen based on real user behavior. Yet in reality, users do not prefer always-on suggestions, meaning that there is something more than text entry speed that is important to the user \cite{quinn2016cost}.

\section{Conclusion and Future Work}
\label{sec:conclusion}

In this paper, we studied the problem of generating inline autocomplete suggestions via sequential decision-making. This formulation allowed us to apply RL methods to solve the problem, and define a reward function that captures user cognitive load through text entry speed.
Our experimental findings suggest that sequential decision-making and RL do \emph{not} improve text entry speed for an idealized user. Therefore, we recommend that further research into sequential decision-making for inline text autocomplete should \emph{not} pursue the goal of solely increasing text entry speed, but rather aim to make real users enjoy the text entry experience.

There are several important directions for future work, to address some of the limitations of this paper. (1) It would be interesting to perform our theoretical analysis on the outputs of a real language model, to connect better with practice. (2) On the computational side, we hope to experiment with feeding the LM probability values as part of the input into the PPO and DQN agents. (3) Our user study only considered prompted writing tasks, where our software told users what sentences to write. The results may be different if users were asked to do freeform writing. (4) Most importantly, we believe that introducing more realistic conditions into our simulations would provide greater scope for RL-based agents to improve over threshold-based methods. Examples include stochasticity in the user suggestion acceptance behavior, typos in their text input, or deciding suggestion acceptance based on semantic matching with the target sentence rather than hard matching. This research would align well with other work that has found users to prefer autocomplete suggestions due to lower cognitive and physical burden, even when text entry speed is impaired~\cite{quinn2016cost}.




\bibliography{main}

\begin{thebibliography}{40}
\providecommand{\natexlab}[1]{#1}
\providecommand{\url}[1]{\texttt{#1}}
\expandafter\ifx\csname urlstyle\endcsname\relax
  \providecommand{\doi}[1]{doi: #1}\else
  \providecommand{\doi}{doi: \begingroup \urlstyle{rm}\Url}\fi

\bibitem[Almeida et~al.(2011)Almeida, Hidalgo, and Yamakami]{almeida2011contributions}
Tiago~A Almeida, Jos{\'e} Mar{\'\i}a~G Hidalgo, and Akebo Yamakami.
\newblock Contributions to the study of sms spam filtering: new collection and results.
\newblock In \emph{Proceedings of the 11th ACM symposium on Document engineering}, pp.\  259--262, 2011.

\bibitem[Arulkumaran et~al.(2017)Arulkumaran, Deisenroth, Brundage, and Bharath]{arulkumaran2017brief}
Kai Arulkumaran, Marc~Peter Deisenroth, Miles Brundage, and Anil~Anthony Bharath.
\newblock A brief survey of deep reinforcement learning.
\newblock \emph{arXiv preprint arXiv:1708.05866}, 2017.

\bibitem[Azzopardi \& Zuccon(2016)Azzopardi and Zuccon]{azzopardi2016analysis}
Leif Azzopardi and Guido Zuccon.
\newblock An analysis of the cost and benefit of search interactions.
\newblock In \emph{Proceedings of the 2016 acm international conference on the theory of information retrieval}, pp.\  59--68, 2016.

\bibitem[Brockman et~al.(2016)Brockman, Cheung, Pettersson, Schneider, Schulman, Tang, and Zaremba]{brockman2016openai}
Greg Brockman, Vicki Cheung, Ludwig Pettersson, Jonas Schneider, John Schulman, Jie Tang, and Wojciech Zaremba.
\newblock Openai gym.
\newblock \emph{arXiv preprint arXiv:1606.01540}, 2016.

\bibitem[Cai et~al.(2016)Cai, De~Rijke, et~al.]{cai2016survey}
Fei Cai, Maarten De~Rijke, et~al.
\newblock A survey of query auto completion in information retrieval.
\newblock \emph{Foundations and Trends{\textregistered} in Information Retrieval}, 10\penalty0 (4):\penalty0 273--363, 2016.

\bibitem[Chen et~al.(2021)Chen, Lu, Rajeswaran, Lee, Grover, Laskin, Abbeel, Srinivas, and Mordatch]{chen2021decision}
Lili Chen, Kevin Lu, Aravind Rajeswaran, Kimin Lee, Aditya Grover, Misha Laskin, Pieter Abbeel, Aravind Srinivas, and Igor Mordatch.
\newblock Decision transformer: Reinforcement learning via sequence modeling.
\newblock \emph{Advances in neural information processing systems}, 34:\penalty0 15084--15097, 2021.

\bibitem[{{CTRL}-labs at Reality Labs} et~al.(2024){{CTRL}-labs at Reality Labs}, Sussillo, Kaifosh, and Reardon]{ctrl2024generic}
{{CTRL}-labs at Reality Labs}, David Sussillo, Patrick Kaifosh, and Thomas Reardon.
\newblock A generic noninvasive neuromotor interface for human-computer interaction.
\newblock \emph{bioRxiv}, pp.\  2024--02, 2024.

\bibitem[Fiorini \& Lu(2018)Fiorini and Lu]{fiorini2018personalized}
Nicolas Fiorini and Zhiyong Lu.
\newblock Personalized neural language models for real-world query auto completion.
\newblock \emph{arXiv preprint arXiv:1804.06439}, 2018.

\bibitem[Fowler et~al.(2015)Fowler, Partridge, Chelba, Bi, Ouyang, and Zhai]{fowler2015effects}
Andrew Fowler, Kurt Partridge, Ciprian Chelba, Xiaojun Bi, Tom Ouyang, and Shumin Zhai.
\newblock Effects of language modeling and its personalization on touchscreen typing performance.
\newblock In \emph{Proceedings of the 33rd annual ACM conference on human factors in computing systems}, pp.\  649--658, 2015.

\bibitem[Gorishniy et~al.(2022)Gorishniy, Rubachev, and Babenko]{gorishniy2022embeddings}
Yury Gorishniy, Ivan Rubachev, and Artem Babenko.
\newblock On embeddings for numerical features in tabular deep learning.
\newblock \emph{Advances in Neural Information Processing Systems}, 35:\penalty0 24991--25004, 2022.

\bibitem[Huang et~al.(2022)Huang, Dossa, Ye, Braga, Chakraborty, Mehta, and Araújo]{huang2022cleanrl}
Shengyi Huang, Rousslan Fernand~Julien Dossa, Chang Ye, Jeff Braga, Dipam Chakraborty, Kinal Mehta, and João~G.M. Araújo.
\newblock Cleanrl: High-quality single-file implementations of deep reinforcement learning algorithms.
\newblock \emph{Journal of Machine Learning Research}, 23\penalty0 (274):\penalty0 1--18, 2022.
\newblock URL \url{http://jmlr.org/papers/v23/21-1342.html}.

\bibitem[Jiang et~al.(2018)Jiang, Chen, Cai, and Chen]{jiang2018neural}
Danyang Jiang, Wanyu Chen, Fei Cai, and Honghui Chen.
\newblock Neural attentive personalization model for query auto-completion.
\newblock In \emph{2018 IEEE 3rd Advanced Information Technology, Electronic and Automation Control Conference (IAEAC)}, pp.\  725--730. IEEE, 2018.

\bibitem[Kaelbling et~al.(1996)Kaelbling, Littman, and Moore]{kaelbling1996reinforcement}
Leslie~Pack Kaelbling, Michael~L Littman, and Andrew~W Moore.
\newblock Reinforcement learning: A survey.
\newblock \emph{Journal of artificial intelligence research}, 4:\penalty0 237--285, 1996.

\bibitem[Kaelbling et~al.(1998)Kaelbling, Littman, and Cassandra]{kaelbling1998planning}
Leslie~Pack Kaelbling, Michael~L Littman, and Anthony~R Cassandra.
\newblock Planning and acting in partially observable stochastic domains.
\newblock \emph{Artificial intelligence}, 101\penalty0 (1-2):\penalty0 99--134, 1998.

\bibitem[Katehakis \& Veinott(1987)Katehakis and Veinott]{bandit1987}
Michael~N Katehakis and Arthur~F Veinott.
\newblock The {Multi-Armed} bandit problem: Decomposition and computation.
\newblock \emph{Mathematics of OR}, 12\penalty0 (2):\penalty0 262--268, May 1987.

\bibitem[Kostrikov et~al.(2021)Kostrikov, Nair, and Levine]{kostrikov2021offline}
Ilya Kostrikov, Ashvin Nair, and Sergey Levine.
\newblock Offline reinforcement learning with implicit q-learning.
\newblock \emph{arXiv preprint arXiv:2110.06169}, 2021.

\bibitem[Lee et~al.(2019)Lee, Hashimoto, and Liang]{lee2019learning}
Mina Lee, Tatsunori~B Hashimoto, and Percy Liang.
\newblock Learning autocomplete systems as a communication game.
\newblock \emph{arXiv preprint arXiv:1911.06964}, 2019.

\bibitem[Levine et~al.(2020)Levine, Kumar, Tucker, and Fu]{levine2020offline}
Sergey Levine, Aviral Kumar, George Tucker, and Justin Fu.
\newblock Offline reinforcement learning: Tutorial, review, and perspectives on open problems.
\newblock \emph{arXiv preprint arXiv:2005.01643}, 2020.

\bibitem[Mitra \& Craswell(2015)Mitra and Craswell]{mitra2015query}
Bhaskar Mitra and Nick Craswell.
\newblock Query auto-completion for rare prefixes.
\newblock In \emph{Proceedings of the 24th ACM international on conference on information and knowledge management}, pp.\  1755--1758, 2015.

\bibitem[Mnih et~al.(2013)Mnih, Kavukcuoglu, Silver, Graves, Antonoglou, Wierstra, and Riedmiller]{mnih2013playing}
Volodymyr Mnih, Koray Kavukcuoglu, David Silver, Alex Graves, Ioannis Antonoglou, Daan Wierstra, and Martin Riedmiller.
\newblock Playing atari with deep reinforcement learning.
\newblock \emph{arXiv preprint arXiv:1312.5602}, 2013.

\bibitem[Murray \& Chiang(2018)Murray and Chiang]{murray2018correcting}
Kenton Murray and David Chiang.
\newblock Correcting length bias in neural machine translation.
\newblock \emph{arXiv preprint arXiv:1808.10006}, 2018.

\bibitem[Paas et~al.(2003)Paas, Tuovinen, Tabbers, and Van~Gerven]{Paas_2003}
Fred Paas, Juhani~E. Tuovinen, Huib Tabbers, and Pascal W.~M. Van~Gerven.
\newblock Cognitive load measurement as a means to advance cognitive load theory.
\newblock \emph{Educational Psychologist}, 38\penalty0 (1):\penalty0 63–71, January 2003.
\newblock ISSN 1532-6985.
\newblock \doi{10.1207/s15326985ep3801_8}.
\newblock URL \url{http://dx.doi.org/10.1207/s15326985ep3801_8}.

\bibitem[Park \& Chiba(2017)Park and Chiba]{park2017neural}
Dae~Hoon Park and Rikio Chiba.
\newblock A neural language model for query auto-completion.
\newblock In \emph{Proceedings of the 40th International ACM SIGIR Conference on Research and Development in Information Retrieval}, pp.\  1189--1192, 2017.

\bibitem[Puterman(2014)]{puterman2014markov}
Martin~L Puterman.
\newblock \emph{Markov decision processes: discrete stochastic dynamic programming}.
\newblock John Wiley \& Sons, 2014.

\bibitem[Quinn \& Zhai(2016)Quinn and Zhai]{quinn2016cost}
Philip Quinn and Shumin Zhai.
\newblock A cost-benefit study of text entry suggestion interaction.
\newblock In \emph{Proceedings of the 2016 CHI conference on human factors in computing systems}, pp.\  83--88, 2016.

\bibitem[Radford et~al.(2019)Radford, Wu, Child, Luan, Amodei, Sutskever, et~al.]{gpt2}
Alec Radford, Jeffrey Wu, Rewon Child, David Luan, Dario Amodei, Ilya Sutskever, et~al.
\newblock Language models are unsupervised multitask learners.
\newblock \emph{OpenAI blog}, 1\penalty0 (8):\penalty0 9, 2019.

\bibitem[Rayner \& Clifton(2009)Rayner and Clifton]{Rayner2009-rm}
Keith Rayner and Charles Clifton, Jr.
\newblock Language processing in reading and speech perception is fast and incremental: implications for event-related potential research.
\newblock \emph{Biol. Psychol.}, 80\penalty0 (1):\penalty0 4--9, January 2009.

\bibitem[Sanh et~al.(2019)Sanh, Debut, Chaumond, and Wolf]{sanh2019distilbert}
Victor Sanh, Lysandre Debut, Julien Chaumond, and Thomas Wolf.
\newblock Distilbert, a distilled version of bert: smaller, faster, cheaper and lighter.
\newblock \emph{arXiv preprint arXiv:1910.01108}, 2019.

\bibitem[Schulman et~al.(2015)Schulman, Levine, Abbeel, Jordan, and Moritz]{schulman2015trust}
John Schulman, Sergey Levine, Pieter Abbeel, Michael Jordan, and Philipp Moritz.
\newblock Trust region policy optimization.
\newblock In \emph{International conference on machine learning}, pp.\  1889--1897. PMLR, 2015.

\bibitem[Schulman et~al.(2017)Schulman, Wolski, Dhariwal, Radford, and Klimov]{schulman2017proximal}
John Schulman, Filip Wolski, Prafulla Dhariwal, Alec Radford, and Oleg Klimov.
\newblock Proximal policy optimization algorithms.
\newblock \emph{arXiv preprint arXiv:1707.06347}, 2017.

\bibitem[Silver et~al.(2016)Silver, Huang, Maddison, Guez, Sifre, Van Den~Driessche, Schrittwieser, Antonoglou, Panneershelvam, Lanctot, et~al.]{silver2016mastering}
David Silver, Aja Huang, Chris~J Maddison, Arthur Guez, Laurent Sifre, George Van Den~Driessche, Julian Schrittwieser, Ioannis Antonoglou, Veda Panneershelvam, Marc Lanctot, et~al.
\newblock Mastering the game of go with deep neural networks and tree search.
\newblock \emph{nature}, 529\penalty0 (7587):\penalty0 484--489, 2016.

\bibitem[Sutton \& Barto(2018)Sutton and Barto]{sutton2018reinforcement}
Richard~S Sutton and Andrew~G Barto.
\newblock \emph{Reinforcement learning: An introduction}.
\newblock MIT press, 2018.

\bibitem[Sutton et~al.(1999)Sutton, McAllester, Singh, and Mansour]{sutton1999policy}
Richard~S Sutton, David McAllester, Satinder Singh, and Yishay Mansour.
\newblock Policy gradient methods for reinforcement learning with function approximation.
\newblock \emph{Advances in neural information processing systems}, 12, 1999.

\bibitem[Tarasov et~al.(2022)Tarasov, Nikulin, Akimov, Kurenkov, and Kolesnikov]{tarasov2022corl}
Denis Tarasov, Alexander Nikulin, Dmitry Akimov, Vladislav Kurenkov, and Sergey Kolesnikov.
\newblock {CORL}: Research-oriented deep offline reinforcement learning library.
\newblock In \emph{3rd Offline RL Workshop: Offline RL as a ''Launchpad''}, 2022.
\newblock URL \url{https://openreview.net/forum?id=SyAS49bBcv}.

\bibitem[Vaswani et~al.(2017)Vaswani, Shazeer, Parmar, Uszkoreit, Jones, Gomez, Kaiser, and Polosukhin]{vaswani2017attention}
Ashish Vaswani, Noam Shazeer, Niki Parmar, Jakob Uszkoreit, Llion Jones, Aidan~N Gomez, {\L}ukasz Kaiser, and Illia Polosukhin.
\newblock Attention is all you need.
\newblock \emph{Advances in neural information processing systems}, 30, 2017.

\bibitem[Volske et~al.(2017)Volske, Potthast, Syed, and Stein]{volske-etal-2017-tl}
Michael Volske, Martin Potthast, Shahbaz Syed, and Benno Stein.
\newblock {TL};{DR}: Mining {R}eddit to learn automatic summarization.
\newblock In \emph{Proceedings of the Workshop on New Frontiers in Summarization}, pp.\  59--63, Copenhagen, Denmark, September 2017. Association for Computational Linguistics.
\newblock \doi{10.18653/v1/W17-4508}.
\newblock URL \url{https://www.aclweb.org/anthology/W17-4508}.

\bibitem[Wang et~al.(2018)Wang, Kolter, Mohan, and Dhillon]{wang2018realtime}
Po-Wei Wang, J~Zico Kolter, Vijai Mohan, and Inderjit~S Dhillon.
\newblock Realtime query completion via deep language models.
\newblock \emph{Proceedings of SIGIR Workshop On eCommerce (SIGIR eCom’18)}, 2018.

\bibitem[Wang et~al.(2020)Wang, Guo, Gao, and Long]{wang2020efficient}
Sida Wang, Weiwei Guo, Huiji Gao, and Bo~Long.
\newblock Efficient neural query auto completion.
\newblock In \emph{Proceedings of the 29th ACM International Conference on Information \& Knowledge Management}, pp.\  2797--2804, 2020.

\bibitem[Wang et~al.(2017)Wang, Ouyang, Deng, and Chang]{wang2017learning}
Yingfei Wang, Hua Ouyang, Hongbo Deng, and Yi~Chang.
\newblock Learning online trends for interactive query auto-completion.
\newblock \emph{IEEE Transactions on Knowledge and Data Engineering}, 29\penalty0 (11):\penalty0 2442--2454, 2017.

\bibitem[Watkins \& Dayan(1992)Watkins and Dayan]{watkins1992q}
Christopher~JCH Watkins and Peter Dayan.
\newblock Q-learning.
\newblock \emph{Machine learning}, 8:\penalty0 279--292, 1992.

\end{thebibliography}
\bibliographystyle{rlc}

\appendix

\section{Reward Derivation}
\label{sec:appendix:reward}
Our objective is to minimize the total time it takes for the user to enter the input. Hence, the instantaneous reward will be proportional to the \emph{saved time} due to a correct suggestion that the user accepted minus the \emph{lost time} due to cognitive load imposed on the user. For simplicity, we assume all characters take the same amount of time to write ($\Delta t_{char-write}$) and to be read ($\Delta t_{char-read}$). We also associate an additional fixed time loss for every suggestion, where the user has to pause and move their gaze ($\Delta t_{distraction}$). Hence we will have:

\begin{align}
r &\propto \text{Saved Time} - \text {Lost Time} \\
&\propto len(a)\times\text{accepted}\times\Delta t_{char-write} - len(a)\times\Delta t_{char-read} - \Delta t_{distraction} \times (len(a) \neq 0)\\
&\propto len(a)\left(\text{accepted} -\frac{\Delta t_{char-read}}{\Delta t_{char-write}}\right) - \frac{\Delta t_{distraction}}{\Delta t_{char-write}}\times (len(a) \neq 0)\\
&\propto len(a)(\text{accepted} - \alpha) - \beta\times (len(a) \neq 0),
\end{align}

where $a$ is the suggestion made by the agent (an empty string if no suggestion was made), and ``accepted'' is a binary indicator for whether the user accepted the suggestion. We estimate $\Delta t_{distraction}$ by two saccada times for the task of natural reading \cite{Rayner2009-rm}, which amounts to $\Delta t_{distraction} = 2 \times 30$ ms = $60$ms. We estimated the per-character writing time and reading time to be $521$ms and $40$ms respectively, based on an internal dataset in our target use case of EMG handwriting~\cite{ctrl2024generic}. This leads to $\alpha = \frac{40}{521}, \beta = \frac{60}{521}$.

\section{Further Details on Computational Experiments}
\label{app:exp_details}
Our PPO and DQN agents are built off the CleanRL implementations~\cite{huang2022cleanrl}, while our IQL agent is built off the CORL implementation~\cite{tarasov2022corl}. For PPO and DQN, we use a per-action $Q$-network architecture that takes as input the state and a single candidate suggestion (or \texttt{wait}), and produces the corresponding $Q$-value (and similarly the action probability for the PPO policy). To obtain the best action, we simply choose the candidate suggestion or \texttt{wait} with highest $Q$-value (or with highest action probability for PPO); note that this requires running $k+1$ forward passes through the model, one per $k$ candidate suggestions plus one for \texttt{wait}. Both PPO and DQN used the \texttt{distilbert-base-uncased} pre-trained model from HuggingFace\footnote{\url{https://huggingface.co/distilbert/distilbert-base-uncased}}, where the Q-function and the PPO policy network are implemented as one-layer projection MLPs on top of the distilbert transformer. We did \emph{not} freeze any layers of the transformer; initial experiments found that only training the projection layer of the Q-function network and using distilbert as a frozen feature extractor worsened DQN's performance. IQL used a multi-layer perceptron (MLP) with two hidden layers of size 256 each, and $k+1$ output nodes, one for each of the $k$ candidates and \texttt{wait}. To obtain the best action, we take the action corresponding to the output node with maximum value. All RL agents were trained for 250K steps.

The default hyperparameters for the RL agents are as follows.

PPO: learning rate $10^{-6}$, number of steps per iteration 20, number of optimization epochs 1, discount factor 0.99. All other hyperparameters were unchanged from the CleanRL defaults.

DQN: learning rate $3 \cdot 10^{-7}$, target network update interval 10000, learning start iteration 25000, discount factor 0.99, model update frequency 10, batch size 32, exploration $\epsilon$ linearly decaying from 1.0 to 0.05 over the first half of training. All other hyperparameters were unchanged from the CleanRL defaults.

IQL: learning rate $3 \cdot 10^{-4}$, batch size 256, discount factor 0.99, $\tau=0.7$. All other hyperparameters were unchanged from the CORL defaults.

\section{Multi-Word Suggestion Length Normalization}
\label{sec:appendix:len_norm}
In our multi-word suggestions experiment, we generated multiple LM candidates via beam search. 
Since the joint likelihood of suggesting two words is always lower than the likelihood of suggesting just the first, i.e., $P(\text{second word} \mid \text{first word}) \times P(\text{first word}) < P(\text{first word})$, we normalized the joint probability of each 2-word suggestion by taking the square root to counteract this effect, similar to \citet{murray2018correcting}.
We note that in the multi-word suggestions experiment, we considered both 1-word and 2-word suggestions, not just the 2-word ones.
Hence, for all suggetions, we universally replace the raw probability from the LM, denoted as $P$, by the normalized probability, denoted as $P'$.
Concretely, the normalized probability $P'$ for a general $m$-word suggestion $w_{1:m}$ after the entered partial sentence $w_{\mathrm{prev}}$ is
\begin{equation*}
    P'(w_{1:m} \mid w_{\mathrm{prev}}) = \left(\prod_{i=1}^m P\left(w_i \mid w_{1:i-1}, w_{\mathrm{prev}}\right) \right)^{\frac{1}{m}}.
\end{equation*}
Here, $w_i$ refers to the $i$th word. We can see that when $m=1$, we simply have $P'(w_1 \mid w_{\mathrm{prev}}) = P(w_1 \mid w_{\mathrm{prev}})$, and hence no length normalization is applied.

Here is a real example from our LM in the multi-word suggestions experiment. For the context \texttt{finally, we have a third val}, the LM generated three candidate completions: \texttt{entine's day}, \texttt{ue}, and \texttt{entine's}. After length normalization, the normalized probabilities were \{\texttt{entine's day}: 33\%, \texttt{ue}: 32\%, \texttt{entine's}: 17\%\}. This would be impossible without length normalization, as the joint probability of \texttt{entine's day} could never exceed the probability of \texttt{entine's}, but it is clear that \texttt{entine's day} is a better completion for the stated context.

\section{Why Does Incorporating 2-Word Suggestions Harm the Oracle?}
\label{sec:appendix:multiword_bad}

In our initial experiments, we were surprised to find that incorporating 2-word suggestions \emph{harms} the performance of even the oracle agent. Recall that the oracle agent has privileged access to the target sentence the user is trying to write, and therefore will never make a wrong suggestion that gets ignored by the user. Upon probing further, we discovered that this is due to several situations where 2-word candidates take up the slot of the correct 1-word candidate, preventing it from being one of the $k$ candidates seen by the agent. Here, we provide a real example from our experiments.

Recall that our 2-word suggestions experiment used $k=5$, i.e., the agent picks from the top-5 candidate completions from the LM with the highest probabilities (together with \texttt{wait}). For the context \texttt{i am gr}, where the user's target sentence is \texttt{i am great, how are you?}, the five candidate completions from our LM, in order of descending probability, are: \texttt{ateful}, \texttt{ateful for}, \texttt{ateful to}, \texttt{eat for}, and \texttt{eat with}. None of these matches the remaining target sentence, so the oracle agent does not surface any suggestion to the user, and instead chooses to \texttt{wait} with a reward of $0$. However, the LM's sixth completion turned out to be \texttt{eat}, which matches the remaining target sentence. Hence, if we were still in the 1-word suggestion scenario, the unmatched 2-word suggestions (\texttt{ateful for}, \texttt{ateful to}, \texttt{eat for}, and \texttt{eat with}) would not be present, and the oracle agent would have received \texttt{eat} as one of its $k=5$ options. The oracle then would have surfaced this suggestion to the user and received positive reward.

\section{Cognitive Load User Study - Extended Results}
\label{sec:appendix:cog-load}

\begin{figure*}[ht]
    \centering
    \includegraphics[width=1\textwidth]{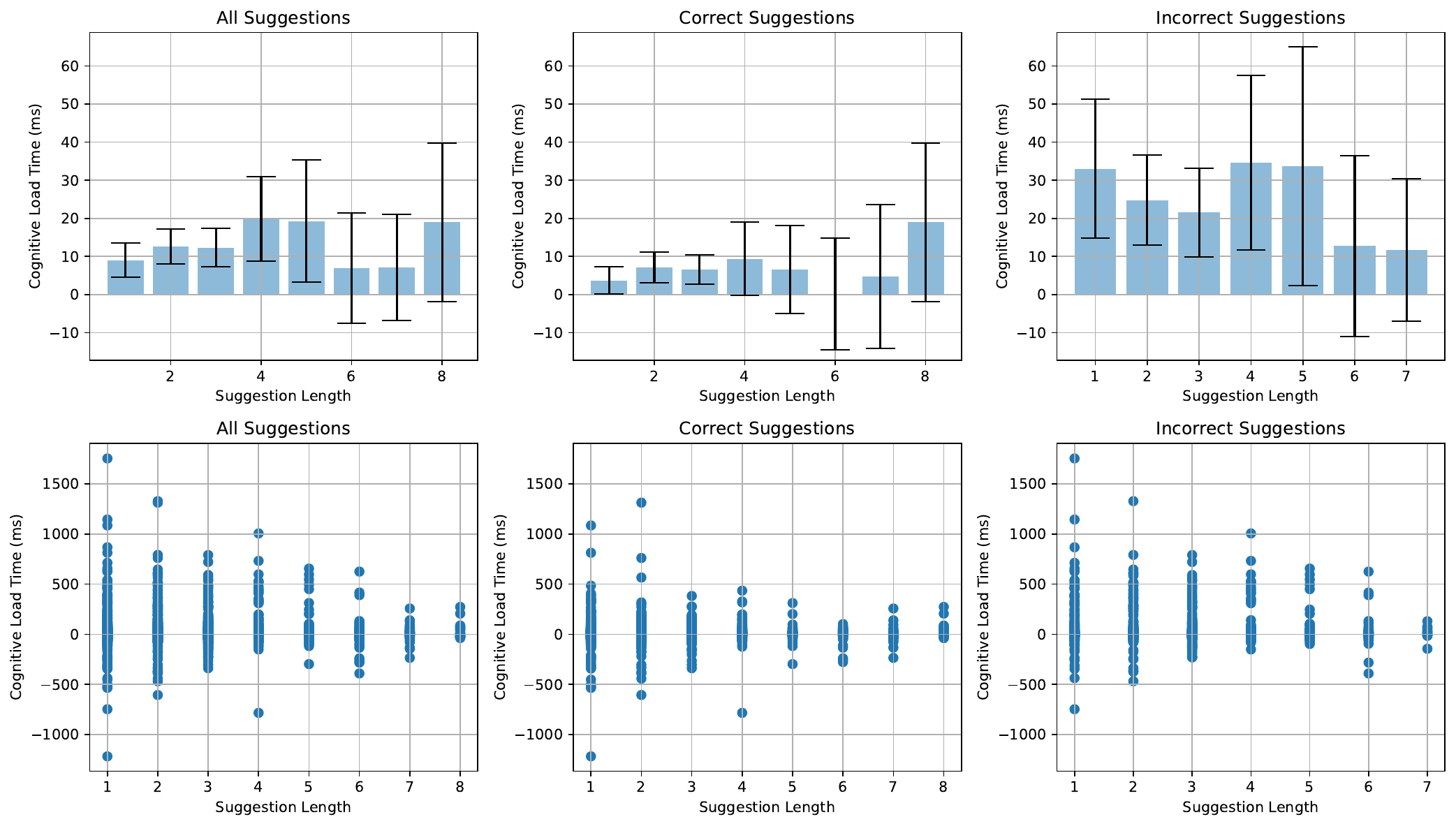}
    \caption{Detailed results of the cognitive load user study based on $N = 9$ subjects. The top row shows the average and 95\% confidence intervals, while the bottom row shows all the actual datapoints. The three columns from left to right represent the average cognitive load across: 1) all suggestions, 2) correct suggestions only, and 3) incorrect suggestions only.}
    \label{fig:appendix:user_study_results}
\end{figure*}

\begin{figure*}[ht]
    \centering
    \includegraphics[width=.75\textwidth]{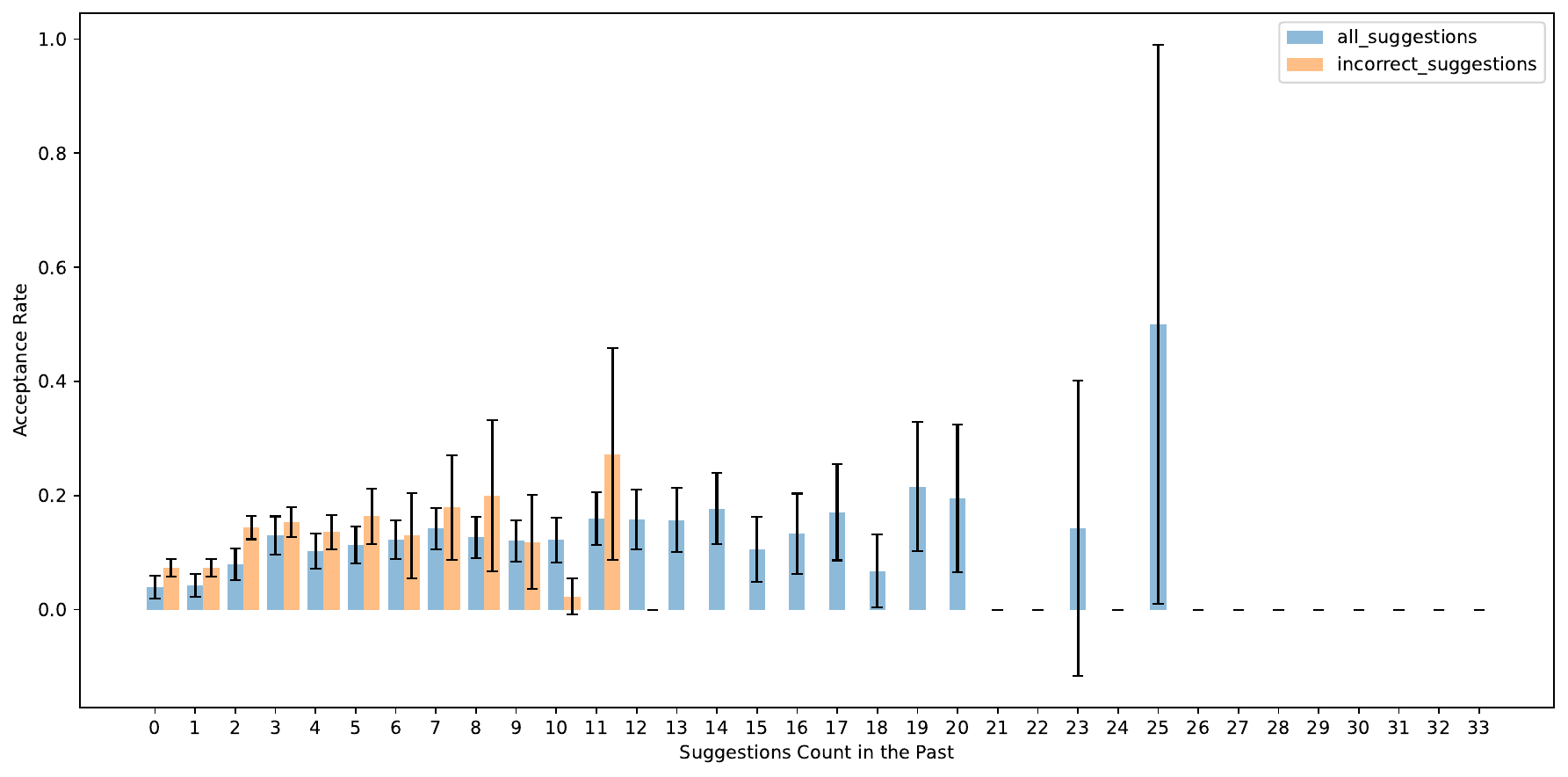}
    \caption{User acceptance rate as a function of the total number of past a) suggestions (blue) and b) incorrect suggestions (orange). The data did not show any evidence of declined suggestion acceptance rate based on the cumulative number of past suggestions, or the cumulative number of past incorrect suggestion. Therefore, we did not find evidence for the accumulated fatigue hypothesis. Error bars depict 95\% confidence interval.}
    \label{fig:appendix:fatigue_results}
\end{figure*}

\end{document}